\documentclass{bmvc2k}

\usepackage{amsfonts}
\usepackage{amsmath}
\usepackage{amssymb}
\usepackage{pifont}
\newcommand{\xmark}{\ding{55}}
\usepackage{comment}
\usepackage{float}
\usepackage{soul} 

\title{Video Summarisation by Classification with Deep Reinforcement Learning}

\addauthor{Kaiyang Zhou}{k.zhou@qmul.ac.uk}{1}
\addauthor{Tao Xiang}{t.xiang@qmul.ac.uk}{1}
\addauthor{Andrea Cavallaro}{a.cavallaro@qmul.ac.uk}{1}

\addinstitution{
Computer Vision Group and \\
Centre for Intelligent Sensing, \\
School of Electronic Engineering and \\
Computer Science, \\
Queen Mary University of London, \\
London E1 4NS, UK
}

\runninghead{Zhou et al.}{Video Summarisation by Classification with Deep RL}

\def\eg{\emph{e.g}\bmvaOneDot}

\def\ie{\emph{i.e}\bmvaOneDot}

\def\vs{\emph{vs}\bmvaOneDot}

\begin{document}

\maketitle

\begin{abstract}
Most existing video summarisation methods are based on either supervised or unsupervised learning. In this paper, we propose a reinforcement learning-based weakly supervised method that exploits easy-to-obtain, video-level category labels and encourages summaries to contain category-related information and maintain category recognisability. Specifically, We formulate video summarisation as a sequential decision-making process and train a summarisation network with deep Q-learning (DQSN).  A companion classification network is also trained to provide rewards for training the DQSN. With the classification network, we develop a global recognisability reward based on the classification result. Critically, a novel {\it dense} ranking-based reward is also proposed in order to cope with the temporally delayed and sparse reward problems for long sequence reinforcement learning. Extensive experiments on two benchmark datasets show that the proposed approach achieves state-of-the-art performance.
\end{abstract}


\section{Introduction}
Video summarisation has traditionally been  formulated as an unsupervised learning problem \cite{wang2016video,wang2017discovering,zhu2013video,zhu2016learning,otani2016video,elhamifar2012see,zhao2014quasi,yang2015unsupervised,zhang2018dtr}, with criteria  to identify keyframes (or key-segments)  hand-crafted based on generic rules, such as diversity and representativeness. However, different types of video content may  require different criteria or different combinations of them: for instance, summaries of {\em Eiffel Tower} videos should contain scenes with the tower, whereas summaries of {\em Making Sandwich} videos should focus on the key temporal stages of the task. How humans deploy these criteria based on the video content can be reflected through their summary annotations, which indicate whether each video frame or segment should be included in the summary. With the annotations, a supervised video summarisation model can be developed \cite{gygli2014creating,zhang2016video,gong2014diverse,gygli2015video,wei2018video,zhao2018hsarnn}, capturing implicitly the content-specific frame/segment selection criteria. However, its use for large-scale summarisation tasks is limited because summary annotations are expensive to collect and prone to bias due to the subjective nature of video summaries.

In this paper, a novel weakly-supervised video summarisation approach is proposed, which is content-specific but only requires video-level annotations in the form of video category labels. These video-level labels are easy to obtain, making the approach much more scalable than the supervised alternatives. Our approach is motivated by the fact that category labels typically encapsulate strong semantic information about the video content. Maintaining the recognisability of the video after removing frames/segments to produce a summary can thus be considered as a top-level selection criterion. Such a criterion encapsulates various fine-grained, content-specific criteria deployed by humans. For example, to summarise videos labelled as {\em Groom Animal}, humans would select segments containing one or more people who are working on an animal to support the semantic meaning conveyed by the category label. We therefore propose to learn a video summarisation model that selects video frames/segments based on whether collectively they contribute the most to recognising the summarised video into its category label.

\begin{figure}[t]
\centering
\includegraphics[width=12.5cm]{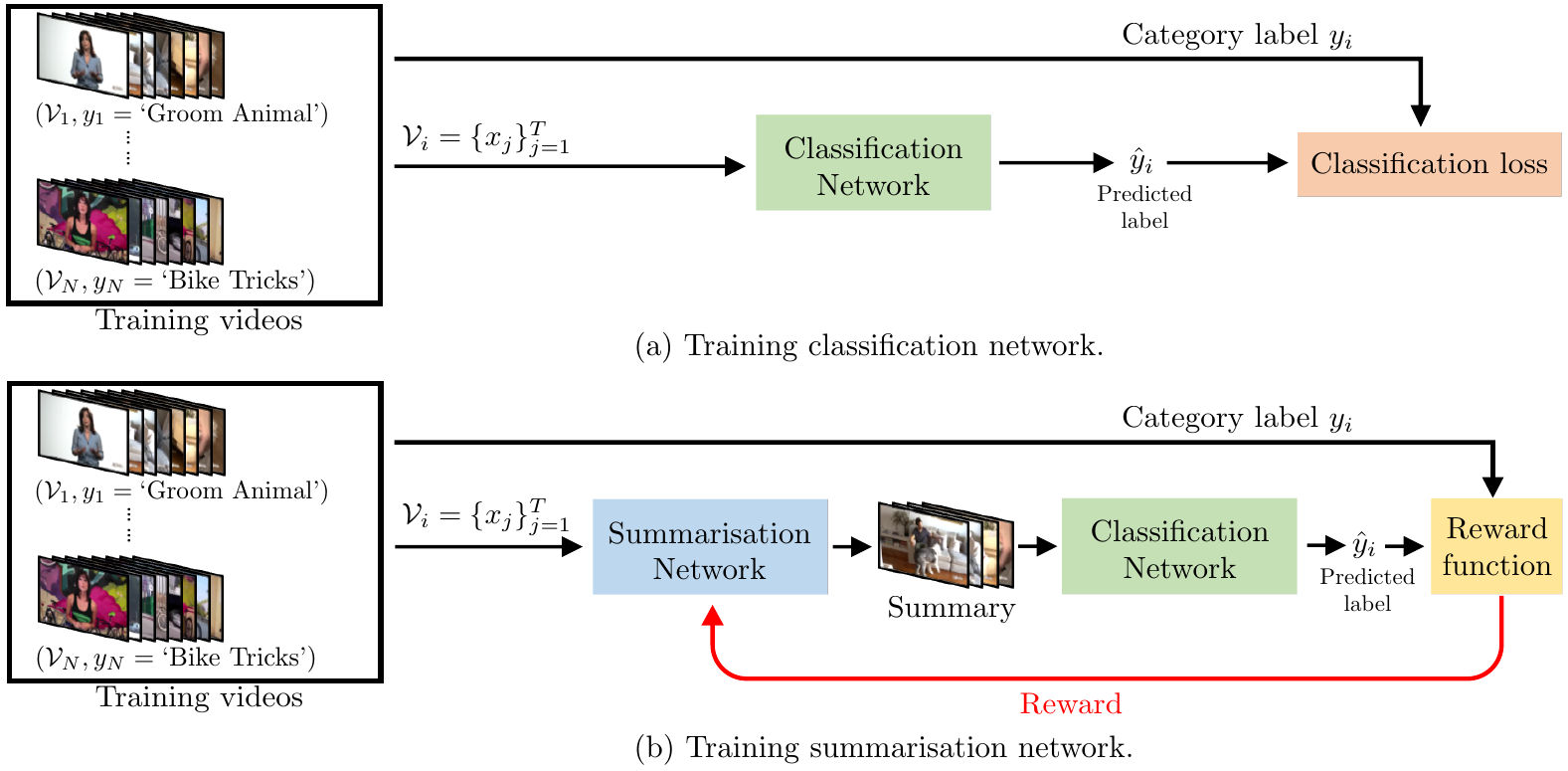}
\caption{Framework overview. After training a classification network with category labels and freezing its weights (a), we train a summarisation network with deep Q-learning with the goal of generating summaries by removing redundant frames, while ensuring  informative parts are maintained and the classification network can recognise them (b).}
\label{fig:overview}
\end{figure}

More specifically, we propose to utilise the video classification criterion elaborated above to guide the learning of a deep video summarisation network. We train the video summarisation network using reinforcement learning (RL) due to the following reasons. First, the classification of summaries can only be made at the end of videos whilst a decision/action needs to be made at every single frame on whether it should be included in the summary. This problem is thus naturally suited for RL. Second, the frame selections are inter-dependent in that the selection of one frame would have implications on the selection of others. The exploration-exploitation strategy of RL can better guide the summarisation network to capture the interdependencies among frames as different combinations of frames are explored.

Figure \ref{fig:overview} shows the proposed framework.  In order to provide rewards during the reinforcement learning of the summarisation network so that the video category recognisability is maintained, our framework includes a companion classification network. This network is a recurrent network learned with supervised classification loss. This classification network  can judge whether a given video sequence contains sufficient information to be classified to a certain category. This judgement is then used as a supervision signal/reward to guide the learning of the summarisation network. Concretely, we formulate the judgement made by the classifier as global recognisability reward and train the summarisation network with deep Q-learning \cite{mnih2013playing,van2016deep}. The summarisation network is thus termed as  Deep Q-learning Summarisation Network (DQSN). Given a video, DQSN generates a summary by sequentially removing frames based on the prediction on future rewards. The classifier then classifies the summary and returns the global recognisability reward to DQSN, which is explicitly encouraged to produce summaries containing category-related information.

A well-known challenge in RL is the credit assignment problem, \ie rewards are sparse or temporally delayed thus making it difficult to associate each action with a reward. With only the global recognisability reward, our DQSN also  suffers from this problem as the single global reward can only be generated after a complete sequence of actions, which inevitably slows down the model convergence. The problem is particularly acute in our case due to the length of the sequences we are dealing with. We mitigate this problem by proposing a novel {\it dense} reward, which we call local relative importance reward. This reward gives each action a feedback by checking if the action changes the recognisability of the partial summary generated so far. Importantly, this reward is also obtained by the classification network, without requiring  additional modules.

{\bf Contributions}.
(1) For the first time, a RL-based weakly supervised video summarisation framework is proposed, which requires only video-level category labels.
(2) We overcome the notorious credit assignment problem in RL by introducing a novel dense reward.
(3) To show the flexibility of our framework, we combine our weakly supervised rewards with those deployed in existing unsupervised approaches and demonstrate their complementarity.
(4) We show that, on two widely-used benchmark datasets, namely TVSum \cite{song2015tvsum} and CoSum \cite{chu2015video},  our approach not only outperforms unsupervised/weakly supervised alternatives but is also highly competitive against supervised approaches.

\section{Related Work}
Existing video summarisation approaches can be categorised as unsupervised, supervised or weakly supervised. Conventional unsupervised approaches cluster frames \cite{wang2016video,wang2017discovering,zhu2013video,zhu2016learning,otani2016video} or optimise hand-crafted objectives \cite{khosla2013large,kim2014joint,song2015tvsum,panda2017collaborative,chu2015video} to identify keyframes or key-segments. The selection criteria are usually generic (\eg diversity \cite{xu2015gaze} and representativeness \cite{elhamifar2012see,zhao2014quasi,yang2015unsupervised}) and do not encode semantics. In contrast, supervised approaches aim to exploit semantics embedded in manually annotated summaries  \cite{gygli2014creating,zhang2016video,gong2014diverse,zhao2018hsarnn,wei2018video,rochan2018learning}. In \cite{lan2018ffnet}, keyframe labels are used to teach neural networks to skip unimportant frames. Since summary annotations are likely to contain biases and are expensive to collect, weak labels such as video category  have been exploited to learn useful concepts to aid summarisation \cite{panda2017weakly,potapov2014category}.

Since our approach is based on weakly-supervised learning of deep network, the  most related video summarisation  work is \cite{panda2017weakly}. In \cite{panda2017weakly},  a 3D ConvNet is trained to predict categories for video clips. Important clips are identified by summing up back-propagated gradients from the true category probability. We significantly improve upon \cite{panda2017weakly} by exploiting category labels with a RL formulation where the interdependencies between frames can be better explored. Our work is also related to \cite{zhou2017deep} in that it is also based on RL. \cite{zhou2017deep}  proposes an unsupervised reward function to train a frame-selection network with policy gradient. Our method differs from \cite{zhou2017deep} in that our global reward function is based on the recognisability of video summaries, while the reward in \cite{zhou2017deep} encourages diversity and representativeness, which are in many cases too generic, as discussed above. Moreover, we use a local dense reward that can compensate the temporally delayed global reward. Our experiments (see Sec.~\ref{sec:experiments}) show that our approach clearly outperforms those in \cite{panda2017weakly,zhou2017deep}.

Beyond video summarisation, several computer vision problems such as image captioning \cite{ren2017deep,wang2017video,zhang2017actor}, visual tracking \cite{yun2017adnet,supancic2017tracking,huang2017learning} and sketch abstraction \cite{umar2018sketchabstract} have been formulated as decision-making processes using RL. A major challenge in RL is credit assignment, which causes difficulties to associating each action with a global sequence-level reward which is sparse and temporally delayed. A common countermeasure is to devise so-called intrinsic rewards, such as the curiosity reward \cite{pathak2017curiosity},  for intermediate states. In our case, the summarisation agent can only receive the reinforcement signal when it finishes a (long) video sequence, thus the reward is severely delayed and sparse. To overcome this issue, we propose a novel dense reward to provide prompt feedback to intermediate states.

\section{Proposed Approach}
Our approach combines two bidirectional recurrent networks with gated recurrent unit (GRU) \cite{cho2014learning}: a classification network (Sec.~\ref{sec:clsnet}) and a summarisation network (Sec.~\ref{sec:sumnet}). Both networks take as input image features extracted by a pretrained ConvNet. We first train the classification network using supervised classification loss and video-level category labels. Then, we apply the fixed classification network to classify the summaries generated by the summarisation network. The classification result is formulated as a reward function and the summarisation network is trained using deep Q-learning \cite{mnih2013playing,van2016deep}. 

\begin{figure}[h]
\centering
\begin{tabular}{cc}
\includegraphics[height=4cm]{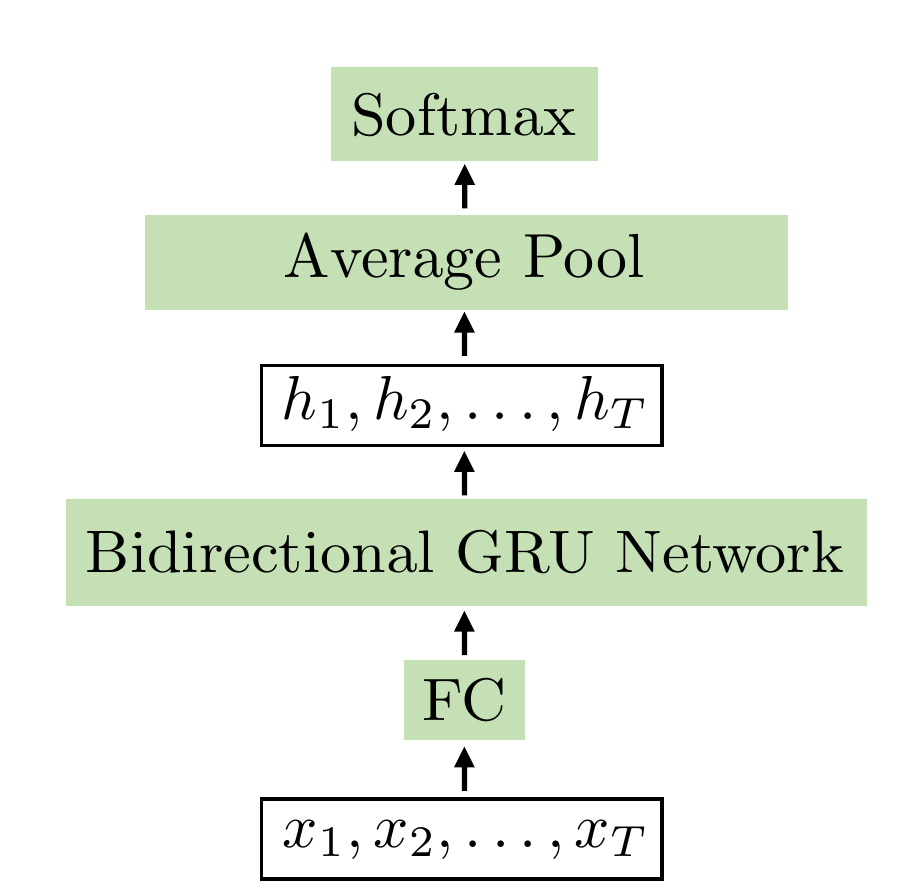} & \includegraphics[height=4cm]{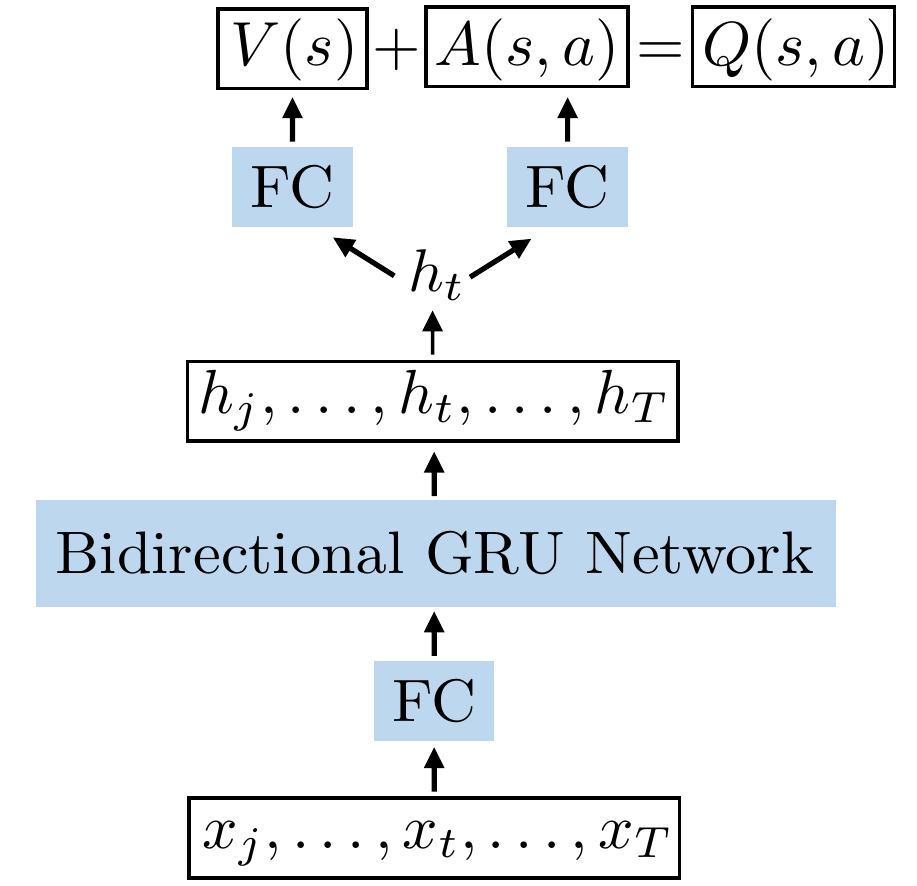} \\
(a) Classification network. & (b) Summarisation network. \\
\end{tabular}
\caption{Network architectures. $x_t$ represents frame features. $h_t$ represents hidden states.}
\label{fig:models}
\end{figure}

\subsection{Classification Network} \label{sec:clsnet}
Fig.~\ref{fig:models}(a) shows the design of our classification network. The input features are first mapped to an embedding space via a fully connected (FC) layer. We use PReLU \cite{he2015delving} as the nonlinear activation function throughout this paper. The embedded features are then processed by a bidirectional GRU (Bi-GRU) network where the outputs from both temporal directions are concatenated, followed by an average pooling layer. Finally, a FC layer with softmax function is mounted on the top to predict $C$ probabilities corresponding to $C$ categories. We train this network using cross entropy loss equipped with the label smoothing regulariser to reduce overfitting \cite{szegedy2016rethinking}. Thus, the loss for a video can be expressed as
\begin{equation}
\mathcal{L} = - \sum_{k=1}^C q (k) \log p (k), \quad s.t. \quad  q (k) = (1 - \omega) \delta(k = y)+ \frac{\omega}{C},
\end{equation}
where $q$ is label, $p$ is prediction, $\delta (\text{condition})$ is 1 if the condition is true otherwise 0, and $\omega$ is weight, which is fixed to 0.1 as suggested in \cite{szegedy2016rethinking}.

\subsection{Deep Q-Learning Summarisation Network} \label{sec:sumnet}
We cast video summarisation as a sequential decision-making process and develop a summarisation network to approximate the action-value function. We term the summarisation network trained with the deep Q-learning algorithm \cite{mnih2013playing} as DQSN (parameterised by $\theta$). From the  reinforcement learning perspective, our framework can be described by a Markov Decision Process (MDP), formally defined as a tuple $(\mathcal{S}, \mathcal{A}, \mathcal{P}, \mathcal{R}, \gamma)$. $\mathcal{S}$ is a set of states experienced by DQSN. A state $s_t$ at time $t$ is defined by a sequence of video frames. $\mathcal{A}$ is the action space composed of two actions: 1 for keeping frame and 0 for discarding frame. $\mathcal{P} (s_{t+1} | s_t, a_t)$ is the transition probability from the current state $s_t$ to the next state $s_{t+1}$ after taking an action $a_t \in \mathcal{A}$. $\mathcal{R} (r_t | s_t, a_t, s_{t+1})$ is the reward for transition $(s_t, a_t, s_{t+1})$. $\gamma \in [0, 1]$ is the discount factor used to reduce the effect of future rewards.

The sequential summarisation process is described as follows. At the first time step $t=1$, the state $s_1$ is composed of the entire sequence of frames of a video, \ie $s_1 = \{ x_j | j = 1, 2, ..., T \}$, but with an attention on $x_1$. DQSN processes $s_1$ and predicts action values $Q_\theta (s_1, a_1)$ for $x_1$. If $Q_\theta (s_1, a_1=1) > Q_\theta (s_1, a_1=0)$, we keep $x_1$ and update next state $s_2 = s_1$. If $Q_\theta (s_1, a_1=1) < Q_\theta (s_1, a_1=0)$, we remove $x_1$ and update $s_2 = s_1 \backslash \{ x_1 \}$. While updating the next state, we simultaneously shift the attention to $x_2$. A reward $r_1$ will be given based on $(s_1, a_1, s_2)$. Iteratively, DQSN processes $s_t$ and predicts action values for frame $x_t$. This process terminates when $t = T$ or the number of remaining frames reaches a threshold.

The objective of DQSN is to take actions that maximise discounted future rewards $R_t = \sum_{t^\prime=t}^T \gamma^{t^\prime-1} r_{t^\prime}$. According to the Bellman equation, DQSN can be trained with deep Q-learning \cite{mnih2013playing} to regress $R_t$:
\begin{equation} \label{eq:dqn}
\mathcal{L}_Q = \mathbb{E}_{s_t, a_t, r_t, s_{t+1}} [ (R_t - Q_\theta (s_t, a_t) )^2 ], \quad s.t. \quad R_t = r_t + \gamma \max_{a_{t+1}} Q_{\theta^-} (s_{t+1}, a_{t+1}),
\end{equation}
where $\theta^-$ represents the parameters of a target network, which is identical to DQSN but is updated periodically. $r_t$ is a hybrid reward, which is detailed in Sec.~\ref{sec:rewardfunc}. In practice, applying a separate network to estimate the future rewards has been proven advantageous to stabilise the training \cite{mnih2013playing}.

Fig.~\ref{fig:models}(b) shows the network architecture of DQSN, whose bottom layers are identical to those of the classification network. The top layers of DQSN aim to predict action values $Q(s, a)$ for frame $x_t$. The bidirectional design allows the past and future information to be jointly captured. Inspired by \cite{wang2016dueling}, we design two streams to produce separate estimates of the value function $V(s)$ and the advantage function $A(s, a)$. $V(s)$ is a scalar that represents the quality of a state achieved by DQSN. $A(s, a)$ is composed of two scalars that provide relative measures of the importance for the two actions. 

Separating $V(s)$ and $A(s, a)$ makes the learning of $Q(s, a)$  more efficient and more robust to numerical scale, as discussed in \cite{wang2016dueling}. To overcome the lack of identifiability between $V(s)$ and $A(s, a)$ when learning $Q(s, a)$, we follow \cite{wang2016dueling} and combine these two functions via
\begin{equation}
Q(s, a) = V(s) + \left(A(s, a) - \frac{1}{| \mathcal{A} |} \sum_{a^\prime} A(s, a^\prime)\right) \quad s.t. \quad \mathcal{A} = \{0, 1\}.
\end{equation}

\subsubsection{Reward Function} \label{sec:rewardfunc}
The reward term $r_t$ in Eq.~\ref{eq:dqn} is a combination of three rewards: global recognisability reward $r_t^g$, local relative importance reward $r_t^l$ and unsupervised reward $r_t^u$.

{\bf Global Recognisability Reward}.
We propose to use the classification results of video summaries as a signal to guide the learning of DQSN. This reward is global because it is only available when a summarisation process finishes, \ie $t = T$. Specifically, if a video summary can be recognised by the classification network, \ie $\hat{y} = y$, we reward the summary with +1, otherwise we penalise the summary with -5\footnote{We give stronger weight to penalty to encourage DQSN to produce summaries with high recognition accuracy, which was found effective in experiments.}. Mathematically, this reward is formulated as
\begin{equation} \label{eq:glbrecognisibility}
r_t^g = \delta(\hat{y} = y) - 5 (1 - \delta(\hat{y} = y)) \quad s.t. \quad t = T.
\end{equation}

{\bf Local Relative Importance Reward}.
To mitigate the credit assignment problem \cite{pathak2017curiosity}, we propose a novel local relative importance reward $r_t^l$, which evaluates the immediate result of removing a frame. The summarisation network can therefore obtain a prompt feedback on the quality of each action. For each transition $(s_t, a_t, s_{t+1})$, the classification network classifies $s_t$ and $s_{t+1}$, resulting in $\xi_t$ and $\xi_{t+1}$, which represent the rank of the true category. For example, if $s_t$ is correctly recognised, $\xi_t = 1$. We introduce relative importance reward based on the change of rank caused by $a_t$. The intuition behind this reward is simple: if the rank is improved, we reward $a_t$; otherwise we penalise $a_t$. To encourage DQSN to remove as many (redundant) frames as possible, we further reward with $+0.05$ for intermediate states if $a_t = 0$. We formulate $r_t^l$ as a function of hyperbolic tangent (tanh):
\begin{equation} \label{eq:localrelativeimportance}
r_t^l = 0.05 (1 - a_t) + h(\xi_t, \xi_{t+1}) \quad s.t. \quad h(\xi_t, \xi_{t+1}) = \text{tanh} \left(\frac{\xi_t - \xi_{t+1}}{\eta}\right), \quad t < T,
\end{equation}
where $\eta$ is a scaling factor. $h(\xi_t, \xi_{t+1})$ measures the importance of $x_t | a_t = 0$ {\it relative} to previously removed frames. Note that this reward is only computed when $a_t = 0$ so it is computationally efficient.

{\bf Unsupervised Rewards}.
Similar to $r_t^g$, $r_t^u$ is also computed globally. We employ the unsupervised diversity-representativeness (DR) reward proposed in \cite{zhou2017deep},
\begin{equation} \label{eq:divrepreward}
r_t^u = \frac{1}{|\mathcal{Y}| |\mathcal{Y} - 1|} \sum_{t \in \mathcal{Y}} \sum_{\substack{t^\prime \in \mathcal{Y} \\ t^\prime \neq t}} d (x_t, x_{t^\prime}) + \exp (- \frac{1}{T} \sum_{t=1}^T \min_{t^\prime \in \mathcal{Y}} || x_t - x_{t^\prime} ||_2),
\end{equation}
where $\mathcal{Y} = \{ y_i | a_{y_i} = 1, i = 1, ..., |\mathcal{Y}| \}$ contains indices of kept frames and $d (\cdot, \cdot)$ is cosine dissimilarity. The first term computes the dissimilarity among selected frames (or segments) while the second term evaluates how well the original video can be reconstructed by the summary. We show in Sec.~\ref{sec:experiments} that the unsupervised DR reward is complementary to our weakly-supervised reward ($r_t^g$ and $r_t^l$). Note that we give equal weights to $r_t^g$ and $r_t^u$.

\subsubsection{Optimisation with Experience Replay and Double Q-Learning}
We employ experience replay \cite{lin1993reinforcement} for minibatch updates. At each time step, we store the experience as a tuple $e_t = (s_t, a_t, r_t, s_{t+1})$ into a replay memory $\mathcal{M}$ initialised with a fixed capacity. 
To update DQSN, we randomly sample minibatches of experiences from $\mathcal{M}$ with uniform distribution. We perform $\epsilon$-greedy policy to select actions, \ie we choose a random action with probability $\epsilon$ and an optimal action from DQSN with probability $1 - \epsilon$.

Q-learning is prone to overestimate action values \cite{hasselt2010double}, as the max operator uses the same function to select as well as to evaluate an action (Eq.~\ref{eq:dqn}). To alleviate this issue, we apply  double Q-learning \cite{hasselt2010double,van2016deep}, its improved version. Specifically, the current Q learner is employed to select the optimal action of the next state and  this action is then evaluated using the target network, so Eq.~\ref{eq:dqn} is substituted with
\begin{equation} \label{eq:dqdn}
\small 
\mathcal{L}_Q = \mathbb{E}_{ \{e_t\} \sim \mathcal{M}} [ (R_t - Q_\theta (s_t, a_t) )^2 ], \quad s.t. \quad R_t = r_t + \gamma Q_{\theta^-} (s_{t+1}, \arg \max_{a_{t+1}} Q_\theta (s_{t+1}, a_{t+1})).
\end{equation}

The gradients are computed based on Eq.~\ref{eq:dqdn}, $\triangledown_\theta \mathcal{L}_Q$. In practice, we replace the squared error loss with Huber loss, which is less sensitive to outliers. To optimise $\theta$, we use Adam \cite{kingma2014adam} and clip the norm of gradients at 5 to avoid exploding gradients \cite{pascanu2013difficulty}.

{\bf Summary Generation}.
During testing, we select actions by $\arg \max_a Q(s, a)$. We score each frame with softmax normalised $Q(s, a=1)$. Shot-level scores are computed by averaging frame scores within the same shots and generate summaries by selecting shots with the highest scores but keeping the duration below a threshold, following \cite{zhang2016video,mahasseniunsupervised,zhou2017deep}. Note that during testing, the video category labels are \textit{not} required.

\section{Experiments} \label{sec:experiments}
We implement our model using PyTorch \cite{paszke2017automatic}\footnote{Code and data will be released at https://github.com/KaiyangZhou.}. We use GoogLeNet \cite{szegedy2015going} trained on ImageNet \cite{deng2009imagenet} to extract frame features followed by $\ell2$ normalisation as the input to the classification and summarisation networks. The dimension of embedding space and hidden units of GRU is  256. The discount factor $\gamma$ is 0.99; $\eta$ in Eq.~\ref{eq:localrelativeimportance} is set to 0.15 via cross-validation; $\epsilon$ in the $\epsilon$-greedy policy decreases exponentially from 1 and stops at 0.1. We set the capacity of $\mathcal{M}$ and minibatch of transitions to 6000 and 200, respectively. The learning rate is $1e-04$.

\subsection{Datasets and Settings}
Experiments are conducted on the widely used TVSum \cite{song2015tvsum} and CoSum \cite{chu2015video} datasets, which contain two sets of non-overlapping categories. TVSum contains 10 categories\footnote{TVSum categories: Changing Vehicle Tire, Getting Vehicle Unstuck, Groom Animal, Making Sandwich, Parkour, Parade, Flash Mob Gathering, BeeKeeping, Bike Tricks, and Dog Show.} each with 5 videos, whose length varies from 2 to 10 minutes. CoSum consists of 51 videos, whose length ranges from 1 to 12 minutes, covering 10 categories\footnote{CoSum categories: Base Jump, Bike Polo, Eiffel Tower, Excavator River Crossing, Kids Playing in Leaves, MLB, NFL, Notre Dame Cathedral, Statue of Liberty, and Surfing.}. Both datasets were annotated by multiple persons so there are multiple human summaries for each video. For evaluation, we compute F-score for each pair of machine summary and human summary and average results for a single video. The overall results are obtained via 5-fold cross-validation, following \cite{zhang2016video,mahasseniunsupervised,zhou2017deep}. We follow \cite{song2015tvsum,chu2015video} to obtain shot-based summaries for evaluation.

\subsection{Training Classification Network}
The training splits of either datasets are not big enough to train our deep classification network well from scratch. Following the existing weakly-supervised summarisation method \cite{panda2017weakly}, we crawl additional video data of the same categories from YouTube for network pretraining. Specifically we search YouTube using the category names as queries. From the returned top-ranked videos, we filter out irrelevant videos using the following rules: (1) length not between 1 and 12 minutes; (2) contain multiple shots; (3) contain dynamic scenes; (4) no cartoons. This leads to 619 videos in total, roughly 30 videos for each category. We call this dataset {\it YouTube619} and use it {\it only} for pretraining the classification network. To test the classification accuracy on YouTube619, we randomly select 100 videos (5 per category) as test set and use the remaining 519 videos as training data to train our Bi-GRU network. We repeat such random split for 5 times and average the test accuracies, obtaining $89.6\%$. We then initialise the network with weights trained on YouTube619 and finetune on the target dataset. As a result, our Bi-GRU classifier with pretraining achieves $74\%$ (TVSum) and $88\%$ (CoSum), outperforming $66\%$ (TVSum) and $72\%$ (CoSum) obtained by the network trained from scratch.

\subsection{Comparison with State-of-the-Art Methods}

\begin{table}[h]
\centering
\begin{tabular}{l|c|c|c}
\hline
Method & Label & TVSum & CoSum \\
\hline
Uniform sampling & \xmark & 15.5 & 20.4 \\
K-medoids & \xmark & 28.8 & 34.3 \\
Dictionary selection \cite{elhamifar2012see} & \xmark & 42.0 & 37.2 \\
Online sparse coding \cite{zhao2014quasi} & \xmark & 46.0 & - \\
Co-archetypal \cite{song2015tvsum} & \xmark & 50.0 & - \\
GAN \cite{mahasseniunsupervised} & \xmark & 51.7 & 44.0 \\
DR-DSN \cite{zhou2017deep} & \xmark & 57.6 & 47.8 \\
\hline 
LSTM \cite{zhang2016video} & frame-level & 54.2 & 46.5 \\
GAN \cite{mahasseniunsupervised} & frame-level & 56.3 & 50.2 \\
DR-DSN \cite{zhou2017deep} & frame-level & 58.1 & \textcolor{red}{54.3} \\
\hline
Backprop-Grad \cite{panda2017weakly} & video-level & 52.7 & 46.2 \\
\hline
DQSN ($r^g$) & video-level & 57.9 & 50.1 \\
DQSN ($r^g + r^u$) & video-level & 58.1 & 51.7 \\
DQSN ($r^g + r^l$) & video-level & \textcolor{blue}{58.2} & 52.0 \\
DQSN (full model) & video-level & \textcolor{red}{58.6} & \textcolor{blue}{52.1} \\
\hline
\end{tabular}
\caption{Summarisation results (\%) on TVSum and CoSum. $1^{st}/2^{nd}$ best in \textcolor{red}{red}/\textcolor{blue}{blue}. Full model means $r^g + r^l + r^u$.}
\label{tbl:cmpwithsoa}
\end{table}

Table \ref{tbl:cmpwithsoa} compares our model, denoted as DQSN (full model), with the  state-of-the-art on TVSum and CoSum. Our findings are summarised as follows.
{\bf (a)} \vs unsupervised: DQSN consistently outperforms all unsupervised approaches, often by large margins. These results suggest that the generic criteria employed in unsupervised learning are limited for not being able to adapt to different types of video content. Among them, the most competitive method is DR-DSN which is also RL-based (with only the generic DR reward). Comparing our DQSN (full model) with DR-DSN, the main difference is on introducing the weakly-supervised rewards (both local and global). Table \ref{tbl:cmpwithsoa} shows that by enforcing recognisability on the generated summaries, our model significantly outperforms DR-DSN. In particular, the improvement margins are $1.0\%$ and $4.3\%$ on TVSum and CoSum, respectively.
{\bf (b)} \vs weakly supervised: We compare DQSN with the recently proposed method based on gradient back-propagation (Backprop-Grad) \cite{panda2017weakly}. It can be seen that our model outperforms Backprop-Grad by $5.9\%$ on both datasets. Both approaches employ recognisability as the frame selection criterion. The superiority of our model over Backprop-Grad can thus be explained by the fact that our RL-based framework can better capture the interdependencies among frames.
{\bf (c)} \vs supervised: Compared to the three supervised methods, LSTM \cite{zhang2016video}, GAN \cite{mahasseniunsupervised} and DR-DSN \cite{zhou2017deep}, all of which require expensive frame-level annotation,  our model is very competitive: it outperforms all three on TVSum; on CoSum, it is only slightly inferior to DR-DSN whilst beating the other two comfortably. Since our approach only uses video-level annotation, it is thus much more suited to large-scale applications.

\subsection{Ablation Study}
In this study we investigate how much different rewards contribute to the final model performance.  Table~\ref{tbl:cmpwithsoa} (bottom rows) compares DQSNs trained with different combinations of rewards (subscript $t$ is omitted). We can see that $r^g + r^l$ clearly outperforms $r^g$ on both datasets, strongly indicating the effectiveness of the local reward. By adding $r^u$, DQSNs are enhanced and achieve better F-scores (while $r^g + r^l + r^u$ still exhibits its advantage over $r^g + r^u$). This thus suggests that the unsupervised DR reward is complementary to our weakly-supervised reward and our approach is flexible enough to incorporate both.

\begin{figure}[h]
\centering
\includegraphics[width=12.5cm]{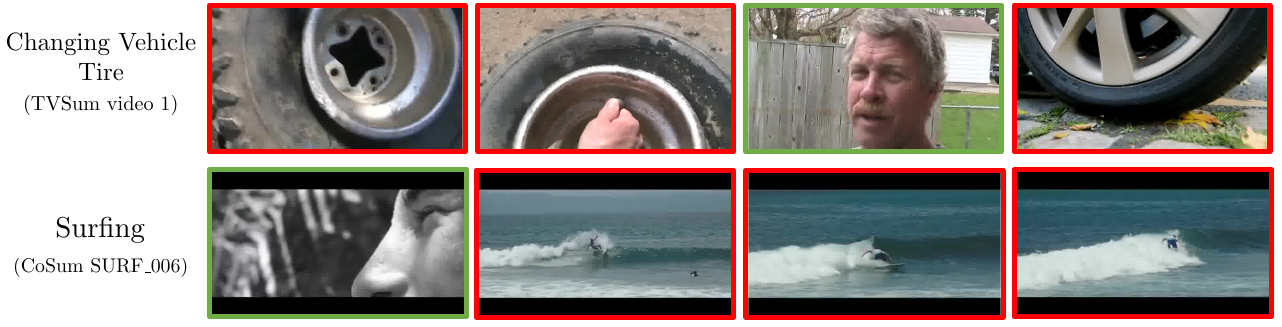}
\caption{Example frames that downgraded (red) / improved (green) the rank of true category in classification when being removed.}
\label{fig:dense_reward_analysis}
\end{figure}

{\bf Why Does Local Relative Importance Reward Help}?
To gain some insights into how the proposed dense local reward contributes, we show some frames that were removed and led to changes of the rank of true category in classification (see Fig.~\ref{fig:dense_reward_analysis}). In the first video, the frames containing the {\em Changing Tire} scene are important because removing them would downgrade the rank of the true category, whereas the frames containing the talking man are relatively unimportant. In the second video, {\em Surfing} frames are apparently more important than `non-surfing' frames. These examples show that local frame relative importance measured by whether it triggers a rank change is indeed a good supervision signal for training the summarisation agent to take the correct action for keeping/removing a given frame.

\subsection{Runtime Efficiency}
We compare the summarisation time of Backprop-Grad \cite{panda2017weakly} and DQSN under the same hardware condition\footnote{We used a GeForce GTX 1080 GPU.} on TVSum + CoSum. Backprop-Grad runs at 3.21 second per video, while DQSN runs at 1.43 second per video, which is more than $2\times$ faster. The reason is because Backprop-Grad performs forward and backward passes on each clip while DQSN only needs to do  a forward pass. Moreover, DQSN consumes less memory as it does not save gradients as Backprop-Grad does.

\subsection{Qualitative Results}
\begin{figure}[h]
\centering
\includegraphics[width=12.5cm]{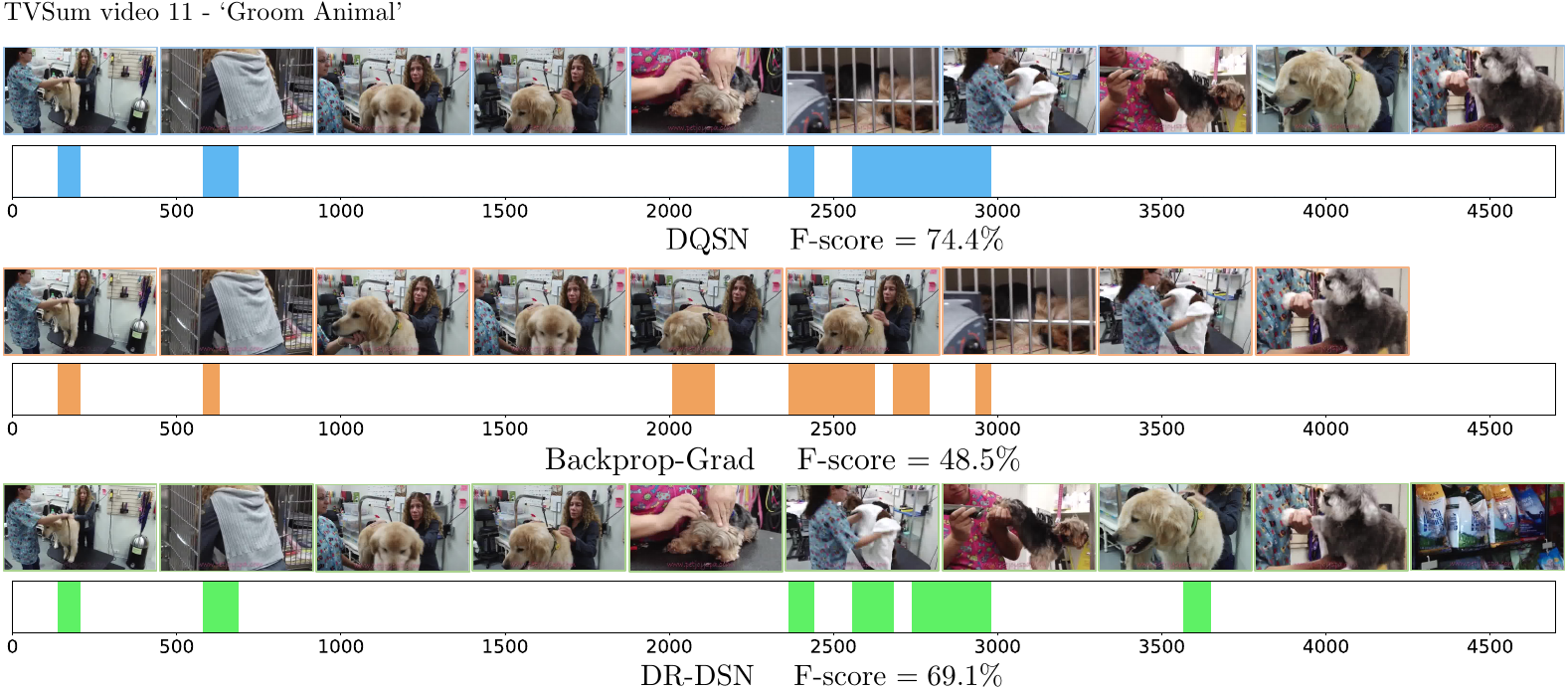}
\caption{Sample summaries obtained by our DQSN, Backprop-Grad and unsupervised DR-DSN. The $x$ axis is the timeline. Coloured segments represent summaries.}
\label{fig:summary}
\end{figure}

Some example summaries are shown in Fig.~\ref{fig:summary}. We observe that DQSN can extract category-related frames containing persons grooming the dogs and  preserve well the temporal storyline. In contrast, Backprop-Grad tends to select repetitive scenes and fails to identify some important details such as the pink-clothed woman grooming the dog. DR-DSN achieves comparable performance to ours, but mistakenly selects irrelevant frames (\eg frames containing dog food) to increase diversity. Such mistake is  due to the fact the generic criteria used in DR-DSN are unable to adapt to the specific video content.

\section{Conclusion}

We presented a RL-based approach DQSN for video summarisation, which uses video-level category labels. A global recognisability reward was formulated to guide the learning of DQSN. Critically, a novel dense reward was proposed to mitigate the credit assignment problem in RL. Compared with unsupervised and supervised learning, our objective function can capture semantics while using only easy-to-obtain, video-level labels.  Experimental results showed that our approach not only outperforms unsupervised/weakly supervised alternatives but is also highly competitive with supervised approaches.

\bibliography{vsummBiB}
\end{document}